\documentclass[11pt,a4paper]{article}

\usepackage[T1]{fontenc}
\usepackage[utf8]{inputenc}
\usepackage[a4paper,margin=2.5cm]{geometry}
\usepackage{graphicx}
\usepackage{booktabs}
\usepackage{multirow}
\usepackage{amsmath}
\usepackage{url}
\usepackage{xcolor}
\usepackage{authblk}
\usepackage{orcidlink}
\usepackage{caption}

\begin{document}

\title{Physiology-Aware Graph Attention for Early Ventricular Reentry Prediction}
\date{}

\author[1]{Ma\"elis Morier\,\orcidlink{0009-0001-2606-0277}}
\author[1]{Jairo Rodr\'iguez Padilla\,\orcidlink{0000-0002-2013-1898}}
\author[4]{Mihaela Pop\,\orcidlink{0000-0003-0474-0324}}
\author[2,3]{Patrick Gallinari\,\orcidlink{0000-0001-9060-9001}}
\author[1]{Maxime Sermesant\,\orcidlink{0000-0002-6256-8350}}

\affil[1]{Epione Research Team, Inria, Universit\'e C\^ote d'Azur, France}
\affil[2]{Sorbonne Universit\'e, CNRS, ISIR, 75005 Paris, France}
\affil[3]{Criteo AI Lab, Paris, France}
\affil[4]{Sunnybrook Research Institute, Toronto, CA}



\maketitle

\begin{abstract}
Sudden cardiac death mostly originates from post-infarct ventricular tachycardia. This reentry is sustained by propagation around scar tissue, but its risk
remains difficult to predict from infarct imaging alone. While \textit{in
silico} induction studies can reveal these mechanisms, exploring many substrates
and pacing locations remains computationally expensive. We propose a graph
attention classifier for early prediction of sustained ventricular reentry from
simulated cardiac electrophysiology. Each simulation is represented as a graph
with early nodal voltage samples and edge attributes encoding geometry,
stimulation frequency, tissue labels, amplitude, and diastolic interval. These
descriptors are computed only from the first two paced activations, before
sustained reentry is observed.  On a balanced independent test set of 1000 simulations, the classifier achieves $94.6$\% accuracy, $95.4$\% sensitivity,
$93.8$\% specificity, $94.6$\% F1-score, $0.987$
ROC-AUC, and $0.985$ PR-AUC. Ablation experiments show that diastolic interval is the most critical descriptor.
Qualitative attention inspection suggests that
different heads focus on border-zone, stimulation location, and early
propagation path. These results support graph attention as a fast
early-prediction approach for large-scale \textit{in silico} ventricular
reentry studies.
\end{abstract}

\section{Introduction}

Ventricular tachycardia (VT) after myocardial infarction is often sustained by
reentrant circuits around scar and border-zone tissue
\cite{colli-franzone_role_2019,donahue_mechanism_2024}. The formation of such
circuits depends on the interaction between scar geometry, local recovery,
slow-conducting tissue, and the stimulation site. Patient-specific
electrophysiology simulations can reproduce pacing protocols and study VT
inducibility \cite{relan_coupled_2011,cedilnik_efficient_2023,trayanova_computational_2024}.
However, testing many scar configurations and pacing locations remains
computationally expensive.

A fast model able to identify simulations likely to develop reentry could
accelerate large-scale \textit{in silico} studies. This is especially useful because autoregressive models can accumulate errors
during long rollouts \cite{bengio_scheduled_2015}, while reentry detection only requires predicting
whether sustained activity will occur.

Graph neural networks (GNNs) are well suited to this problem because cardiac meshes are naturally represented as graphs \cite{meister_graph_2020}. Previous graph-based approaches have
mainly focused on activation-map estimation or surrogate modeling of cardiac
electrical propagation
\cite{meister_graph_2020,morier_learning_2025,martinez_full-field_2026}.
Other machine-learning methods have addressed arrhythmia classification from
ECG signals, rotor localization, or prediction of action potential traces
\cite{zhang_deep_2025,lebert_rotor_2021,mcgillivray_machine_2018,shahi_long-time_2021}.
In contrast, we study early reentry prediction directly on irregular cardiac
meshes.

Our proposed approach is a physiology-aware graph attention classifier for early
prediction of sustained ventricular reentry. Each simulation is represented as a graph, with nodes containing early voltage samples, and edge attributes describing geometry and physiological information. The model predicts whether the simulation will later develop sustained reentry, without observing the reentry phase itself. Our contributions are:
\begin{enumerate}
\item A graph attention classifier for early prediction of sustained ventricular
reentry from simulated cardiac electrophysiology.
\item An ablation study showing the role of diastolic interval, tissue labels,
geometry, and early voltage information.
\item An exploratory inspection of attention maps as qualitative explanations.
\end{enumerate}

\begin{figure}[h!]
    \centering
    \includegraphics[width=0.96\textwidth]{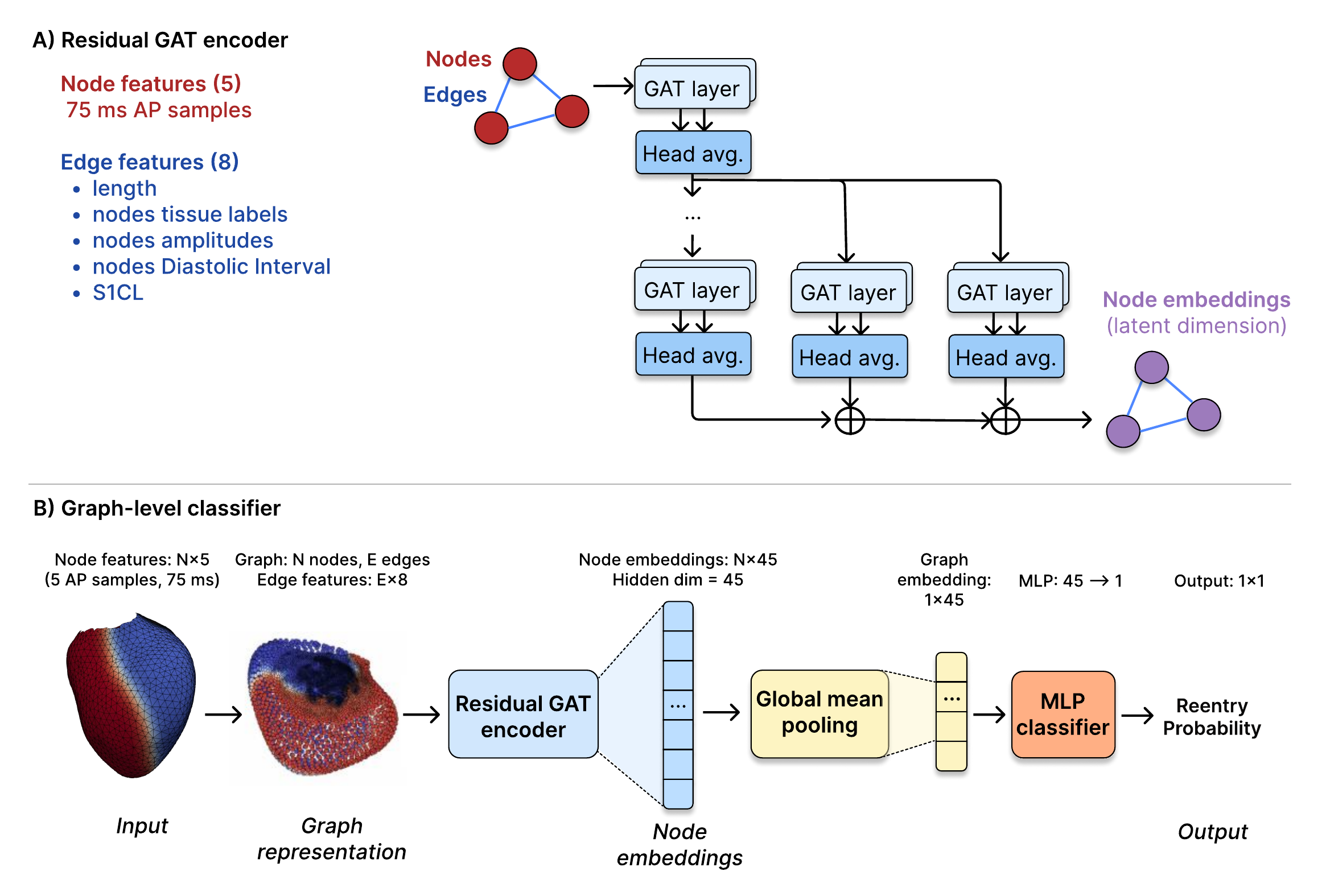}
\caption{
Overview of the proposed graph classifier. A residual GAT encoder produces node embeddings, which are averaged into a graph-level representation and passed to an MLP classifier to predict sustained reentry.
}
    \label{fig:classifier_framework}
\end{figure}

\section{Methods}
Figure~\ref{fig:classifier_framework} summarizes the full pipeline. We first
generate simulated cardiac graphs from ventricular meshes with scar and
border-zone regions. Each graph contains early AP samples as node features and
physiological descriptors as edge features. A residual GAT encoder then maps the
graph to node embeddings, which are pooled and classified to predict sustained
reentry.

\subsection{Electrophysiology Simulations}

\paragraph{Synthetic simulations.}
We generated ventricular action-potential (AP) simulations using a modified
Mitchell--Schaeffer (MS) model \cite{mitchell_two-current_2003}, which describes the
normalized transmembrane potential $u \in [0,1]$ and a recovery variable $h\in [0,1]$:
\begin{equation}
\begin{aligned}
\partial_t u &=
\operatorname{div}(\mathbf{D}\nabla u)
+ \frac{h u(u-\lambda)(u_{\max}-u)}{\tau_{\mathrm{in}}}
- \frac{u}{\tau_{\mathrm{out}}}
+ J_{\mathrm{stim}}, \\
\partial_t h &=
\begin{cases}
(1-h)/\tau_{\mathrm{open}}, & u \leq u_{\mathrm{gate}}, \\
-h/\tau_{\mathrm{close}}, & u > u_{\mathrm{gate}} .
\end{cases}
\end{aligned}
\end{equation}
Here, $\mathbf{D}$ is the diffusion tensor and $J_{\mathrm{stim}}$ is the
external stimulus. Further model details are given in
\cite{djabella_two-variable_2007}. All simulations used the same MS ionic
parameter setting. Parameter values for healthy tissue were:
$\tau_{\mathrm{in}}=0.3$, $\tau_{\mathrm{out}}=6$,
$\tau_{\mathrm{open}}=120$, $\tau_{\mathrm{close}}=150$,
$u_{\mathrm{gate}}=0.13$, $u_{\max}=1$, $\lambda=0.01$, and $D=0.15$.
For border-zone tissue:
$\tau_{\mathrm{in}}=0.3$, $\tau_{\mathrm{out}}=6$,
$\tau_{\mathrm{open}}=120$, $\tau_{\mathrm{close}}=265$,
$u_{\mathrm{gate}}=0.13$, $u_{\max}=0.9$, $\lambda=0.14$, and $D=0.03$.
Scar tissue used the border-zone time constants and gate threshold, with
$u_{\max}=0$ and $\lambda=0.25$. The pacing cycle length was
$S1CL=310$. All time constants and $S1CL$ are in ms, except $u_{max}$, $u_{gate}$ and $\lambda$ (dimensionless).
The meshes were derived from \cite{escartin_extent_2023,ly_interpretable_2022}.
For each mesh, scar and border-zone regions were generated procedurally by sampling patch centers, assigning nearby nodes as border-zone tissue, and embedding smaller low-conductivity scar islands. Simulations were run for 6.1 s using an implicit--explicit time-integration scheme \cite{ascher1995implicit,rodriguezpadilla2025}.

\paragraph{Dataset splitting.}
The dataset was split by original anatomy to evaluate generalization to unseen
meshes. The training set contains simulations from 12 meshes, the validation set
from 4 meshes, and the test set from 2 held-out meshes. For each mesh, multiple
scar/border-zone configurations and stimulation locations were simulated. Thus,
the same anatomical substrate could produce either reentry or non-reentry
depending on the pacing site. This design reduces the risk of solving the task
by scar recognition alone.

The dataset contains 5000 training simulations, 2004 validation simulations,
and a balanced independent test set of 1000 simulations. The training and validation sets were imbalanced, with reentry cases
representing roughly 15--22\% of the simulations, reflecting the lower
prevalence of sustained reentry in the generated dataset. In contrast, the
independent test set was balanced with 500 cases per class to evaluate
sensitivity and specificity under controlled conditions.

\paragraph{Reentry labels.}
Labels were assigned from the full simulation outcome. A simulation was labelled as reentry when organized reentrant activity persisted
until $t=6$ s, and as non-reentry otherwise. We used this threshold as a practical compromise between computational and
clinical criteria for sustained ventricular tachycardia
\cite{campos_automated_2022,ng_virtual_2012}.
As a sensitivity check, we recomputed labels on the test set using earlier
late-activity thresholds of 4.0, 5.0, and 5.5 s.

\paragraph{Graph representation.}
Each simulation mesh was represented as a graph, with mesh vertices as nodes and
mesh adjacency as edges. Node features contained the first five saved voltage samples, corresponding to
75 ms of transmembrane voltage. This short window captures the beginning of the paced activation, before
sustained reentry develops. Edge attributes encoded edge length, tissue labels,
stimulation frequency, and node-wise amplitude and diastolic interval (DI)
values for the two incident nodes. Amplitude and DI were computed only from the
first two paced activations, before sustained reentry was observed. DI was
defined as the interval between repolarization after the first paced activation
and the next paced activation. The stimulation frequency was kept for compatibility with future variable pacing
protocols, although it was fixed in the present dataset and therefore did not
carry discriminative information here. Together, these features provide pre-reentry information about local geometry, tissue organization, and recovery timing.

\subsection{Proposed Model}

\paragraph{Graph Classifier.}
The proposed classifier takes as input the cardiac graph described above. We use
an edge-conditioned GATv2 encoder \cite{velickovic_graph_2018,brody_how_2022},
so that message passing can depend on both the local AP state and the edge
descriptors. 

The encoder contains six GATv2 layers with two attention heads per layer and a
hidden dimension of 45. The final two layers include additive residual
connections from the first hidden representation, following the common use of
residual connections to improve information flow in deeper neural networks
\cite{he_deep_2016}. The node embeddings are aggregated with global mean pooling
and passed to an MLP that outputs one logit for graph-level reentry
classification.

\paragraph{Training details.}
Models were trained on the training split, with checkpoints selected on the
validation dataset and final performance reported only on the independent test
set. To account for class imbalance during training, we used class-weighted
binary cross-entropy. The positive-class weight was set to the non-reentry /
reentry ratio. We used AdamW with learning rate $10^{-4}$, weight decay
$10^{-6}$, batch size 8, and gradient clipping with maximum norm 1.0. Models
were trained for up to 1000 epochs, while retaining the checkpoint with the
highest validation F1-score. The learning rate was reduced when the validation
F1-score plateaued (factor 0.5, patience 15 epochs, minimum learning rate
$10^{-6}$). A fixed threshold of 0.5 was used to classify test simulations as
reentry or non-reentry. This threshold was used as a fixed operating point for
sensitivity and specificity, while ROC-AUC and PR-AUC were reported as
threshold-independent metrics. 

\paragraph{Non-graph baselines.}
We also evaluated two non-graph baselines. The SVM-RBF baseline used one
26-dimensional global summary vector per simulation. The NodeMLP baseline used
node-level voltage, amplitude, DI, and tissue inputs, followed by mean/max
pooling, but no graph message passing.

\section{Experiments and Results}
\subsection{Classification Performance}

\paragraph{Balanced evaluation.}
Across five independent random seeds, performance was stable on the balanced
anatomy-disjoint test set, with $93.94\pm0.49$\% accuracy and
$93.98\pm0.44$\% F1-score. We report the seed selected by validation F1 as the
main model. It achieved 94.6\% accuracy, 93.9\% precision, 95.4\% sensitivity,
93.8\% specificity, 94.6\% F1-score, 0.987 ROC-AUC, and 0.985 PR-AUC, with 477
true positives, 469 true negatives, 31 false positives, and 23 false negatives.
This indicates that sustained
reentry can be predicted with high sensitivity from early activation
information and graph-level physiological descriptors, while maintaining good
specificity.

\paragraph{Threshold and calibration.}
The threshold 0.5 was used as a fixed operating point, while ROC-AUC and PR-AUC
were reported as threshold-independent metrics. The Brier score was 0.0505 and
the 10-bin expected calibration error was 0.0312. Because the model was trained
with class-weighted loss on synthetic data, probabilities are interpreted as
screening scores rather than calibrated clinical risks. Recomputing the
sustained-reentry label at 5.0 and 5.5 s gave identical labels to the original
6.0 s criterion, and using 4.0 s changed only one test label.

\paragraph{Imbalanced evaluation.}
We also evaluated the same trained model on an imbalanced subset of the
independent test set, using the same held-out meshes but a lower reentry
prevalence of approximately 20\%. This proportion was chosen within the range of
reported clinical inducibility rates \cite{soris_programmed_2025,deshmukh_duration_2020}.
The same decision threshold was used, without additional tuning. On this subset,
the model achieved 94.6\% accuracy, 93.2\% specificity, and 88.1\% F1-score,
with 63 true positives, 233 true negatives, 17 false positives, and no false
negatives. This corresponds to 100.0\% sensitivity on the 63 reentry cases, which
should be interpreted with caution given the smaller number of positives.
Precision decreased to 78.8\%, as expected under lower reentry prevalence.

\begin{figure}[h]
\centering
\includegraphics[width=0.61\textwidth]{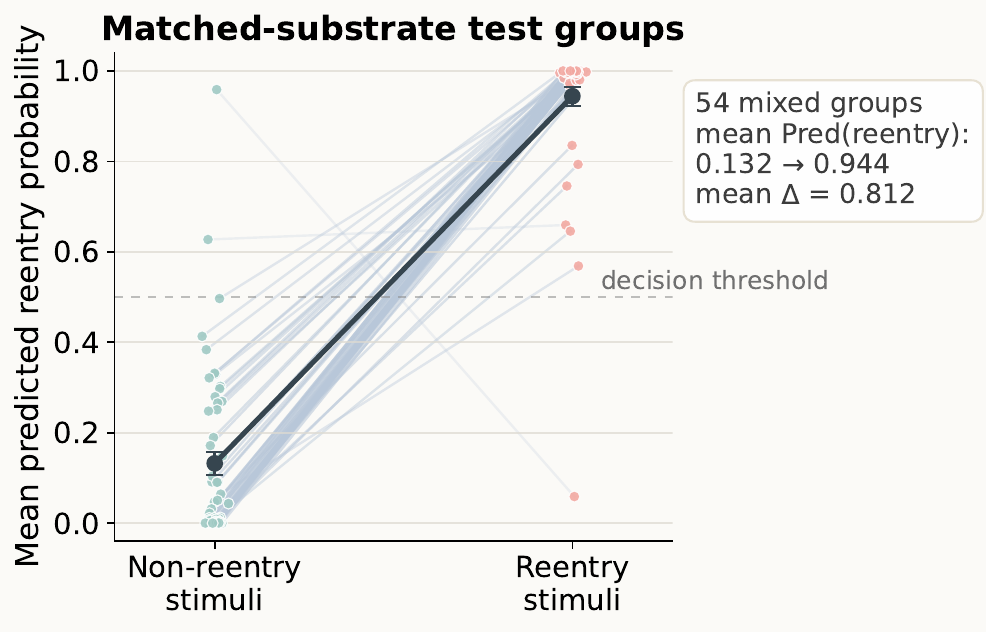}
\caption{
Matched-substrate analysis. Lines connect mean reentry probabilities for
non-reentry and reentry stimuli with the same substrate.
}
\label{fig:matched_substrate}
\end{figure}

\paragraph{Matched-substrate evaluation.} To verify that the model was not simply recognizing scar patterns, we analyzed
test cases with the same mesh and scar/border-zone configuration but different
stimulation locations, producing both reentry and non-reentry propagation.
Among 113 matched-substrate groups in the test set, 54 contained both reentry
and non-reentry outcomes. When averaging predictions within each matched-substrate group, the mean
predicted probability increased from 0.132 for non-reentry stimuli to 0.944 for reentry stimuli (Fig.~\ref{fig:matched_substrate}).

\subsection{Ablation Study and Baselines}

Table~\ref{tab:ablations} summarizes the ablation experiments on the balanced
test set. 

\paragraph{Architectural baselines.}
The full model achieved high sensitivity while maintaining good specificity.
We compared it with two architectural baselines: a sequential GAT
with the same number of GATv2 layers but without residual additions, and a GCN
baseline using the same input features and number of layers. The sequential GAT
remained competitive, but showed lower sensitivity. The GCN baseline detected
most reentry cases, but its low specificity indicates that it misclassified many
non-reentry simulations as reentry. This suggests that residual GATv2 message
passing with edge-conditioned attention provides a better trade-off between
detecting reentry and rejecting non-reentry cases. The SVM-RBF baseline uses
global graph-summary features, while the NodeMLP uses node-level features but no
message passing.

\begin{table}[h!]
\centering
\caption{Ablation results on the balanced independent test set.}
\label{tab:ablations}
\scriptsize
\begin{tabular}{p{0.35\textwidth}ccccc}
\toprule
Model variant & Accuracy & Precision & Sensitivity & Specificity & F1  \\
\midrule
\multicolumn{6}{l}{\textit{Architectural and non-graph baselines}} \\
\textbf{Proposed residual GAT} & \textbf{0.9460} & 0.9390 & \textbf{0.9540} & 0.9380 & \textbf{0.9464} \\
Sequential GAT & 0.9240 & 0.9398 & 0.9060 & 0.9420 & 0.9226 \\
GCN baseline & 0.6310 & 0.5796 & 0.9540 & 0.3080 & 0.7211 \\
SVM-RBF, global summaries & 0.8590 & 0.9860 & 0.7280 & \textbf{0.9900} & 0.8380 \\
NodeMLP, no message passing & 0.9130 & 0.9210 & 0.9040 & 0.9220 & 0.9120 \\

\midrule
\multicolumn{6}{l}{\textit{Input-window and feature ablations}} \\
Short voltage window (2 samples) & 0.9370 & 0.9524 & 0.9200 & 0.9540 & 0.9359 \\
Edge attributes only & 0.9120 & 0.9682 & 0.8520 & 0.9720 & 0.9064 \\
Voltage only & 0.6330 & 0.5893 & 0.8780 & 0.3880 & 0.7052 \\
Tissue tags only & 0.5000 & 0.5000 & 1.0000 & 0.0000 & 0.6667 \\
DI only & 0.8990 & 0.9716 & 0.8220 & 0.9760 & 0.8906 \\
Amplitude only & 0.5550 & 0.6608 & 0.2260 & 0.8840 & 0.3368 \\

\midrule
\multicolumn{6}{l}{\textit{Feature-drop ablations}} \\
No edge length & 0.9010 & 0.9740 & 0.8240 & 0.9780 & 0.8927 \\
No amplitude & 0.9280 & 0.9693 & 0.8840 & 0.9720 & 0.9247 \\
No DI & 0.6440 & 0.6417 & 0.6520 & 0.6360 & 0.6468 \\
No tissue tags & 0.9090 & 0.9745 & 0.8400 & 0.9780 & 0.9023 \\
\bottomrule
\end{tabular}
\end{table}

\paragraph{Effect of the voltage window.}
The short-window ablation preserved high performance with only two voltage
samples, showing that the result is not driven by a finely tuned 75 ms input
window. We kept the five-sample window in the proposed model because it was
defined a priori as a short early observation period and achieved the best
accuracy, sensitivity, and F1-score.

\paragraph{Input-window and feature ablations.}
Input and feature ablations showed that edge attributes carried most of the
predictive signal. Edge attributes alone reached 91.2\% accuracy and 90.6\%
F1-score, whereas voltage-only inputs were much less specific. Tissue tags alone
predicted all cases as reentry, confirming that scar and border-zone labels are
insufficient on their own. DI alone remained predictive, whereas amplitude alone
missed many reentry cases.

\paragraph{Feature-drop ablations.}
Feature-drop ablations further confirmed the importance of DI. Removing DI
caused the largest degradation, with an F1-score of 0.6468 and lower
sensitivity and specificity. This effect is physiologically
plausible: DI summarizes local recovery timing after the early paced
activations, indicating which regions are excitable, partially recovered, or
still refractory when the next wavefront arrives. Since reentry depends on the
combination of delayed propagation and heterogeneous recovery around scar and
border-zone tissue, DI provides a key descriptor of the vulnerable substrate.

\paragraph{Overall interpretation.}
Removing amplitude, edge length, or tissue tags caused smaller performance
drops than removing DI. Overall, these results indicate that the model does not
rely on scar geometry alone. Instead, it combines recovery timing, geometry,
amplitude, tissue labels, and early voltage information to obtain a balanced
prediction.

\subsection{Computational Cost}

To quantify the potential benefit of early screening, we compared the runtime of
the full simulation pipeline with the inference time of the trained classifier.
The full pipeline, including Mitchell--Schaeffer simulation and conversion to
graph format, required $1076.6\pm215.8$ s per simulated case on CPU. Classifier inference was measured on the same CPU architecture. Forward
inference on the 1000 test graphs required $11.35\pm0.06$ ms per graph. This
corresponds to an approximate forward-inference speed-up of
$9.5\times10^4$ compared with the full simulation pipeline. Including graph
loading, model-side evaluation remained approximately $5.0\times10^4$ faster. This comparison should be interpreted as an order-of-magnitude runtime estimate
rather than a low-level solver benchmark, but it supports the use of the
classifier as a fast screening step before running full electrophysiology
simulations.

\subsection{Model Inspection}

We inspected attention maps to see which regions were highlighted by the classifier before graph-level pooling. This analysis is qualitative only.

\begin{figure}[h]
\centering
\includegraphics[width=\textwidth]{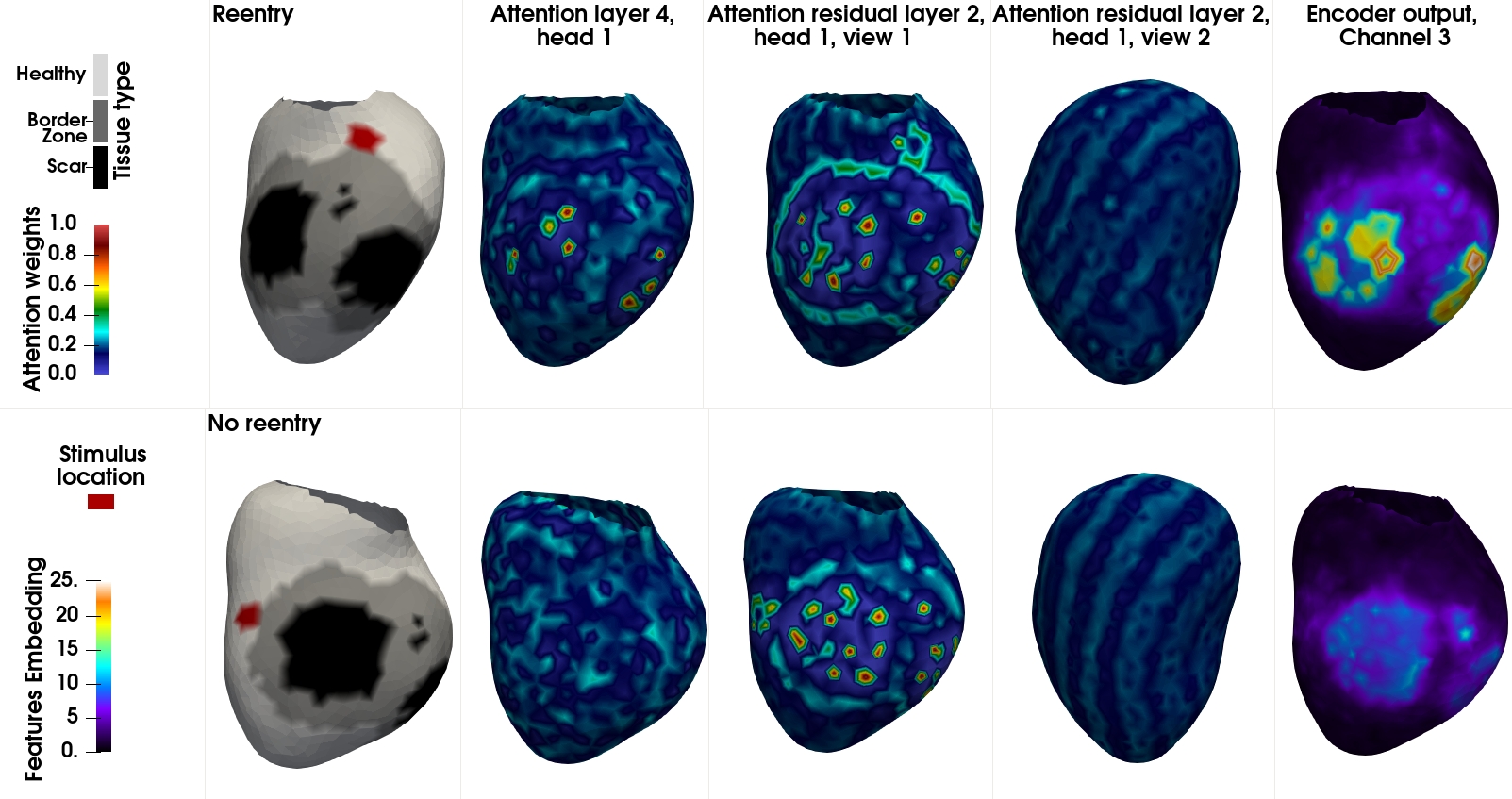}
\caption{
Selected attention maps and one encoder-output channel for a matched reentry
and non-reentry pair.  
}
\label{fig:attention_layers}
\end{figure}

\paragraph{Qualitative attention maps.}
Figure~\ref{fig:attention_layers} shows selected maps for a matched reentry and
non-reentry pair. The first column shows tissue type and stimulation site. The
next columns show attention maps from layer 4, head 1, and from the last
residual layer in two different views. The last column shows channel 3 of the
final node embedding before graph-level pooling. Layer 4, head 1 shows stronger localized hotspots in the reentry case than in
the matched non-reentry case. A similar difference is visible in embedding
channel 3, suggesting that this local attention signal may be retained in the
final node representation. The residual-layer maps are more global, highlighting
broader regions related to the stimulus location, scar and border-zone shape,
and early propagation. Together, these maps suggest that the classifier combines localized hotspots
with broader spatial information before graph-level pooling. We do not claim
that these regions identify the critical isthmus or the reentry pathway. They
should be interpreted as visual cues that may be interesting to inspect further. Additional propagation videos and attention-map examples are provided in the
supplementary material.

\paragraph{Scar-associated hotspot analysis.}
We also analyzed layer 4, head 1 on the full balanced test set. We counted a
case as having a strong scar hotspot when at least one scar node had attention
above a chosen threshold. This pattern was strongly associated with reentry. At
threshold 0.8, a strong scar hotspot was present in 454/500 reentry cases
(90.8\%), but only in 68/500 non-reentry cases (13.6\%). The separation was
stable across thresholds from 0.6 to 0.9: strong scar hotspots were present in
88.4--93.0\% of reentry cases, compared with 8.2--21.4\% of non-reentry cases.
This suggests that layer 4, head 1 captures a scar-associated signal that is
much more often activated in reentry simulations. These hotspots may therefore indicate regions worth inspecting further, but
their physiological meaning remains to be validated.

\section{Limitations}

This study uses synthetic Mitchell--Schaeffer simulations with one global ionic
parameter setting and a fixed stimulation protocol. This design isolates the
effects of geometry, tissue organization, and stimulation location, but future
work should test broader electrophysiological parameters, cycle lengths, pacing
protocols, realistic scar morphologies, and patient-derived substrates. The
train, validation, and test splits were anatomy-disjoint, but the balanced test
set contained only two held-out anatomies; validation on larger and more diverse
patient cohorts is therefore required. The classifier should also be
interpreted as a fast simulation-screening tool, not as a calibrated clinical
risk predictor, especially because training used class-weighted loss on
synthetic data. The classifier predicts whether sustained reentry will occur,
but does not yet localize the reentry pathway or identify an ablation target.
Future work should combine early reentry prediction with localization tasks,
such as critical isthmus or late-activation pathway detection. Finally, the
attention maps are exploratory. They should be interpreted as visual cues.
Expert review or a dedicated localization benchmark would be needed to determine
whether these regions are physiologically meaningful.

\section{Conclusion}

We presented a graph-attention classifier for early prediction of sustained
ventricular reentry in simulated post-infarct substrates. The model predicts
future reentry from early voltage samples and physiological graph descriptors
with high sensitivity and specificity. Ablation experiments show that DI is a
key descriptor, but that the best performance is obtained by combining recovery
timing with geometry, tissue labels, and early voltage information.
Matched-substrate examples further suggest that the model captures
pacing-dependent changes in reentry inducibility rather than scar geometry
alone, and non-graph baselines indicate that graph message passing improves over
using global or node-wise descriptors without propagation on the mesh. CPU
timing suggests that the classifier can screen candidates orders of magnitude
faster than running the full simulation pipeline. This approach could therefore
reduce computational cost by selecting which pacing locations or substrates
should be prioritized for full simulation, while future work should validate the
method on more diverse and clinically realistic substrates.

\section*{Acknowledgments}
This work has received funding from the French government through the National
Research Agency (ANR) PEPR Digital Health ChroniCardio project (2023000289), the
ANR DeepNum project (ANR-21-CE23-0017), and through the ANR 3IA Côte d’Azur
and IA Cluster project (ANR-19-3IA-0002 and ANR-23-IACL-0001). This work was
granted access to the HPC resources of IDRIS under the allocation
AD011015902R1 made by GENCI. The authors are grateful to the OPAL infrastructure
from Universit\'e C\^ote d'Azur for providing resources and support.

Some of the cardiac geometries used in this study were obtained from data generated with support from the Canadian Institutes of Health Research (CIHR), Project Grant PJT-153212.

\section*{Disclosure of Interests}
The authors have no competing interests to declare that are relevant to the
content of this article.

\bibliographystyle{plain}
\bibliography{paper_morier_stacom}
\end{document}